\documentclass[letterpaper]{article}

\usepackage{amsmath}
\usepackage{amssymb}
\usepackage{natbib,alife}  
\usepackage{url,hyperref,cleveref}
\usepackage{booktabs}
\usepackage{algorithm}
\usepackage{algpseudocode}
\usepackage{subfig}
\usepackage{float}
\usepackage{xcolor}

\newcommand{\G}{\mathcal{G}}
\newcommand{\Real}{\mathbb{R}}

\newcommand{\git}{\url{https://github.com/erwanplantec/LNDP}}

\title{Evolving Self-Assembling Neural Networks: From Spontaneous Activity to Experience-Dependent Learning} 

\author{
   Erwan Plantec,
    Joachim W. Pedersen, 
       Milton L. Montero,
    Eleni Nisioti, \and 
    Sebastian Risi \\
    \mbox{}\\
    IT University of Copenhagen, Denmark \\
    \{erpl, sebr\}@itu.dk
} 

\begin{document}

\maketitle

\begin{abstract}
Biological neural networks are characterized by their high degree of plasticity, a core property that enables the remarkable adaptability of natural organisms. Importantly, this ability affects both the synaptic strength and the topology of the nervous systems. Artificial neural networks, on the other hand, have been mainly designed as static, fully connected structures that can be notoriously brittle in the face of changing environments and novel inputs. Building on previous works on Neural Developmental Programs (NDPs), we propose a class of self-organizing neural networks capable of synaptic and structural plasticity in an activity and reward-dependent manner which we call \emph{Lifelong Neural Developmental Program} (LNDP). We present an instance of such a network built on the graph transformer architecture and propose a mechanism for pre-experience plasticity based on the spontaneous activity of sensory neurons. We demonstrate the model's ability to learn from experiences in different control tasks starting from randomly connected or empty networks. We further show that structural plasticity is advantageous in environments necessitating fast adaptation or with non-stationary rewards.
\end{abstract}

\section{Introduction}

Millions of years of open-ended evolutionary processes have led to the emergence of complex cognitive abilities, which are primarily supported by brains composed of continually adapting neurons and synapses. These highly plastic biological neural networks enable organisms to quickly adapt in ontogenetic timescales through learning, a major evolutionary transition in the evolution of cognition \citep{ginsburg2021}. Furthermore, by growing through a temporally extended developmental phase from a DNA representation into an extremely complex system, brains also exhibit a high degree of adaptability on a phylogenetic timescale. 

On the other hand, artificial neural networks (ANNs) have been mainly built as fixed fully connected structures that remain frozen after training, unable to adapt to any unexpected change. Indeed, even though much research has been devoted to the topic of online and meta-learning \cite{hochreiter2001,andrychowicz2016,wang2018c}, it is still the case that the most modern neural network systems use offline learning as it is much simpler to use in combination with backpropagation.

Fundamentally, biological and artificial neural networks seem to follow diametrically opposed design philosophies: self-organization and engineering \citep{ha2022a}. In the past decade, self-organization has gained a lot of interest in the AI community as a path towards closer to natural artificial intelligence \citep{risi2021selfassemblingAI}. Several works have attempted to design ANNs endowed with more of the characteristics found in living organisms in order to replicate properties such as fast adaptability and lifelong learning, or generalization capabilities \citep{kudithipudi2022,parisi2019}. For instance, Evolved Plastic Artificial Neural Networks (EPANNs) have been proposed as a research framework aiming at building such neural networks through neuroevolution \citep{soltoggio2018a}, while work on indirect encodings of artificial neural networks has proposed several models of self-assembling neural networks \citep{najarro2022b} (see \citet{stanley2003a} for a taxonomy of developmental encodings for artificial neural networks). 

Recently, \cite{najarro2023} proposed a model capable of assembling functional artificial neural networks represented as graphs, which they call Neural Developmental Programs (NDPs). However, these models are temporally limited to a pre-environmental phase and do not account for lifetime, and even less lifelong learning.

In this work, we propose a way to address those limitations by extending the NDP framework. Specifically, we incorporate a mechanism that enables plasticity and structural changes throughout the lifetime of an agent. It achieves this by performing local computations which depend on both the local activity of each neuron in the ANN and the global reward function from the environment. The resulting system thus defines a family of plastic neural networks bridging the gap between indirect developmental encodings and meta-learned plasticity rules which we call \emph{Lifelong Neural Developmental programs} (LNDP). 

LNDPs are defined by a set of possibly parametric components defining neural and synaptic dynamics. Motivated by works on architectural priors \citep{bhattasali2022a} and the role of structural plasticity in learning \citep{caroni2012,may2011}, LNDPs are also endowed with structural plasticity (i.e.\ synapses can be dynamically added or removed). We further propose an instance of an LNDP based on a Graph Transformer layer \citep{dwivedi2021}, modeling neuronal communication through synaptic and extra-synaptic channels, which enables neurons to self-organize and differentiate.  Neural and synaptic dynamics are then modeled with Gated Recurrent Units (GRUs) \citep{chung2014}. These components together define a large family of learnable LNDPs which we can optimize to solve different reinforcement learning tasks. 

Motivated by works exploring the role of spontaneous activity (SA) in developing biological \citep{luhmann2016,leighton2016,kirkby2013a} and artificial \citep{raghavan2019a,bednar1998,valsalam2007} neural networks, we further extend the system with a mechanism that enables pre-experience development based on SA. Here we model SA with a simple learnable stochastic process of sensory neurons. Using SA enables the reuse of the same components during both pre-experience development and experience-dependent learning.

In this work, we demonstrate the effectiveness of the LNDP in driving artificial neural networks to self-organize in an activity and experience-dependent manner into functional networks able to solve control tasks, starting from randomly connected or empty networks. We further present evidence that structural plasticity improves results in environments requiring fast adaptation or with non-stationary dynamics requiring continuous adaptation. Additionally, the effectiveness of pre-environmental spontaneous activity-driven developmental phases to drive the network's self-organization into functional units is demonstrated.

\section{Background}
This section reviews relevant related works, such as meta-learning synaptic plasticity, developmental encodings, and approaches that investigate the intersection between structural and synaptic plasticity.
\subsection{Meta-learning Synaptic Plasticity}
Multiple works have tried to meta-learn synaptic plasticity rules to discover new learning rules for ANNs potentially enabling faster and more sample efficient adaptation of artificial agents. For instance, several works have meta-trained gradientless synaptic plasticity rules enabling faster and more sample-efficient learning than traditional stochastic gradient descent \citep{sandler2021b,kirsch2022}. Recently, \citeauthor{pedersen2024} have proposed a model called Structurally Flexible Neural Networks (SFNNs), which are able to learn from rewards in different environments with different input and output dimensions \citep{pedersen2024}. More biologically inspired self-organized networks, based on Spike Timing-Dependent Plasticity-like rules have also been developed for reservoir computing \citep{lazar2009}. \citeauthor{aswolinskiy2015} also proposed a reward-modulated version of the same networks for reinforcement learning \citep{aswolinskiy2015}.  \cite{najarro2020} evolved Hebbian synaptic plasticity rules allowing agents to perform well in locomotion tasks from initially random networks and even adapt to damages unseen during training.

\subsection{Developmental Encodings}
Developmental encodings are a subset of indirect encodings characterized by a process unrolling in time while indirect encodings can be direct ``one-shot" mappings (like hypernetworks \citep{ha2022b}). For instance, L-systems are able to grow   artificial neural networks through a process of development/rewriting   \citep{campos2011}. In \citet{lowell2016}, the authors used a model of gene regulatory networks for this purpose and showed it led to the emergence of modular networks. Neural Cellular Automata (NCA) have also been demonstrated to allow the control of neural network development \citep{najarro2022b}. More recently, \cite{najarro2023} proposed NDPs which are instances of developmental encodings based on graph neural networks.

\section{Lifelong Neural Developmental Programs} \label{sec:model}
On a high level, the Neural Developmental Program (NDP) introduced in \cite{najarro2023}, defines a set of neural components that can be optimized to direct the growth of an artificial neural network into functional networks in a fully decentralized manner (i.e.\ neurons perceive the state of the neural network through a graph neural network layer and can independently of each other decide to divide). However, the approach was so far restricted to growing neural structures that were fixed during the lifetime of the agent. This section presents the Lifelong Learning NDP (LNDP) approach, which enables learning to modify the weights and the connectivity of the neural network while the agent interacts with the environment. This ability is especially important for environments with non-stationary rewards that require lifetime learning. 
A similar objective has been addressed in \citep{ferigo2023}. However, in their approach, the number of trainable parameters of their model is tied to the size of the produced networks while this is not the case for NDPs. 

In more detail, we represent a neural network at a given time $t$ as a directed graph $\G^t$ with $N$ nodes (neurons) which is fully described by the tuple $<A^t, I, O, h^t, e^t, v^t, w^t>$. $A^t$ is the $N \times N$ binary adjacency matrix indicating the presence of an edge in between two nodes. $I$ and $O$ are the sets of input and output nodes respectively. $h^t \in H^{N}$ and $e^t \in E^{N^2}$ are the nodes and edges states respectively with $H \equiv \Real^{dh}$ and $E \equiv \Real^{de}$. Edge states are masked by the adjacency matrix, i.e.\ set to zero wherever the adjacency matrix is 0. $v^t \in \Real^N$ are the nodes' activations and $w^t$ is the real-valued weight matrix. We refer as $h_i$ to the state of node $i$ and $e_{ij}$ to the state of the edge between nodes $i$ and $j$.  

In the following, we describe the components that together make a LNDP. These components can be parameterised learnable functions. We note the full set of parameters as $\theta$.

\paragraph{Initialization}
We note $\G^0 \equiv <A^0, I, O, h^0, e^0, v^0, w^0>$ as the initial network, which is sampled from a distribution $\Psi = \mathbb{P}(\G^0=g)$. Input and output neurons stay fixed during the agent's lifetime.

\paragraph{Nodes}
$f_{\theta}^h: \G \to H^N$ the node function, which updates the nodes' states based on the complete graph state, is defined as: 
\begin{equation}
    h^{t+1} = f_{\theta}^h(\G^t).
\end{equation}
Using the complete graph state to update the nodes enables taking information from the network structure through interactions between nodes (e.g with graph neural networks). While many different types of information can be used in the node update, taking activity states $v^t$ into account allows for activity-dependent mechanisms. Intuitively, one can see the nodes' states as the molecular or membrane states of neurons which can, in turn, affect plasticity and neuronal dynamics. 

\paragraph{Edges}
Edges' states are updated through a function $f_{\theta}^e: H \times H \times \Real \to E$, which updates their states based on the states of connected nodes and a reward signal $r^t$ received from the environment. We refer to this function as the edge or synapse model:
\begin{equation}
    e_{ij}^{t+1} = f_{\theta}^e(e_ij^{t}, h^{t+1}_i, h^{t+1}_j, r^t).
\end{equation}
Following,  a weight model function $f_{\theta}^w: E \to \Real$  associates a scalar weight to each edge based on its state. Note that if nodes are updated using activities, then some edge update functions could be able to learn activity-dependent rules such as Hebbian learning: 
\begin{equation}
    w_{ij}^t = f^w_{\theta}(e^t_{ij}).
\end{equation}

\paragraph{Structural plasticity}

Structural plasticity in the network is achieved by allowing new edges to be added (i.e.\ synaptogenesis) and existing edges to be removed (i.e.\ pruning). 
$f_{\theta}^+: H^2 \to [0,1]$ is the synaptogenesis function which defines the probability of adding an edge (synapse) between two nodes at each time step:
\begin{equation}
    \mathbb{P}(A_{ij} \gets 1) = f_{\theta}^+(h_i^t, h_j^t). 
\end{equation}

Similarly, $f_{\theta}^-: E \to [0, 1]$ is the pruning function and defines the probability of removing an edge in the graph (i.e.\ setting its entry in the adjacency matrix to $0$) as a function of its state: 
\begin{equation}
    \mathbb{P}(A_{ij}\gets 0) = f_{\theta}^-(e_{ij}^{t}).
\end{equation}

\paragraph{Network dynamics}

We define the network dynamics through a dynamical model $\phi$ updating the nodes' activations based on past activations, the current weight matrix $w^t$, nodes' states $h^t$, and observation $o^t$: 
\begin{equation}
    v^{t+1} = \phi(v^t, w^t, h^t, o^t) 
\end{equation}
Node states can be used to define neuron parameters such as biases.

\subsection{Implementation}

\begin{figure}[ht]
    \centering
    \subfloat[Node model]{\includegraphics[width=0.8\columnwidth]{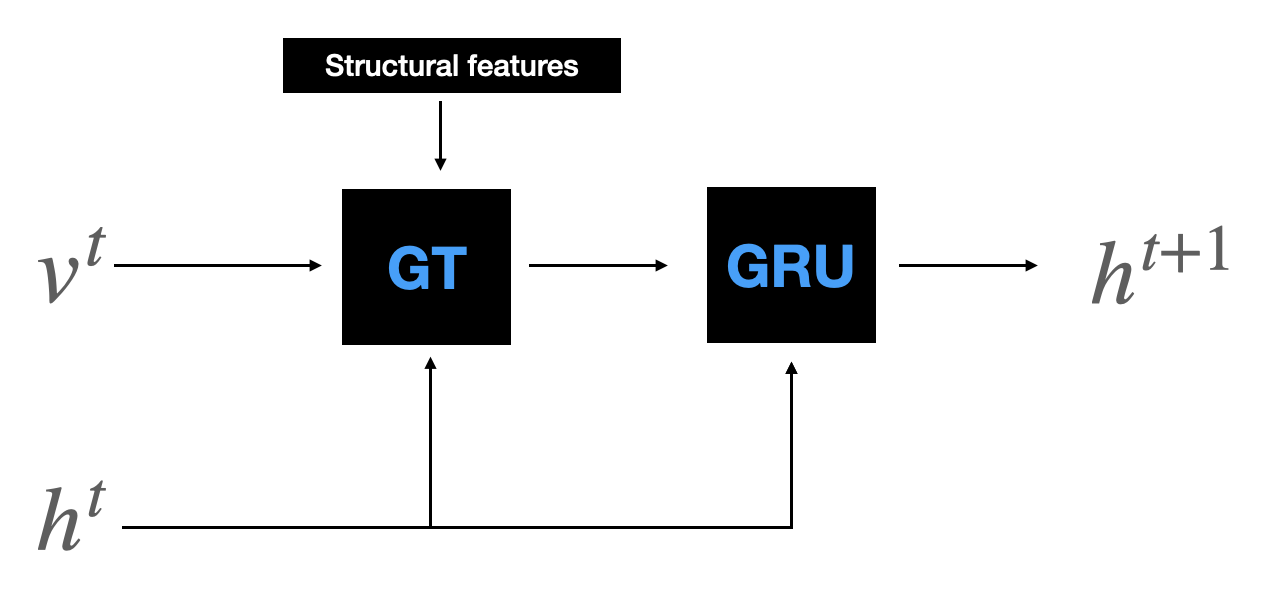}}\\
    \subfloat[Edge model]{\includegraphics[width=.5\columnwidth]{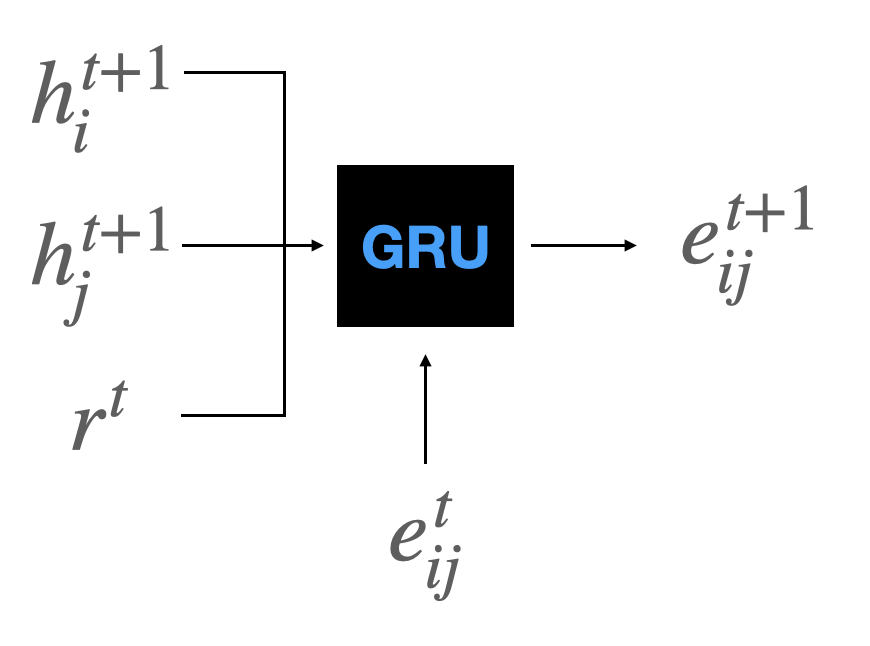}}
    \subfloat[Network topology]{\includegraphics[width=0.5\columnwidth]{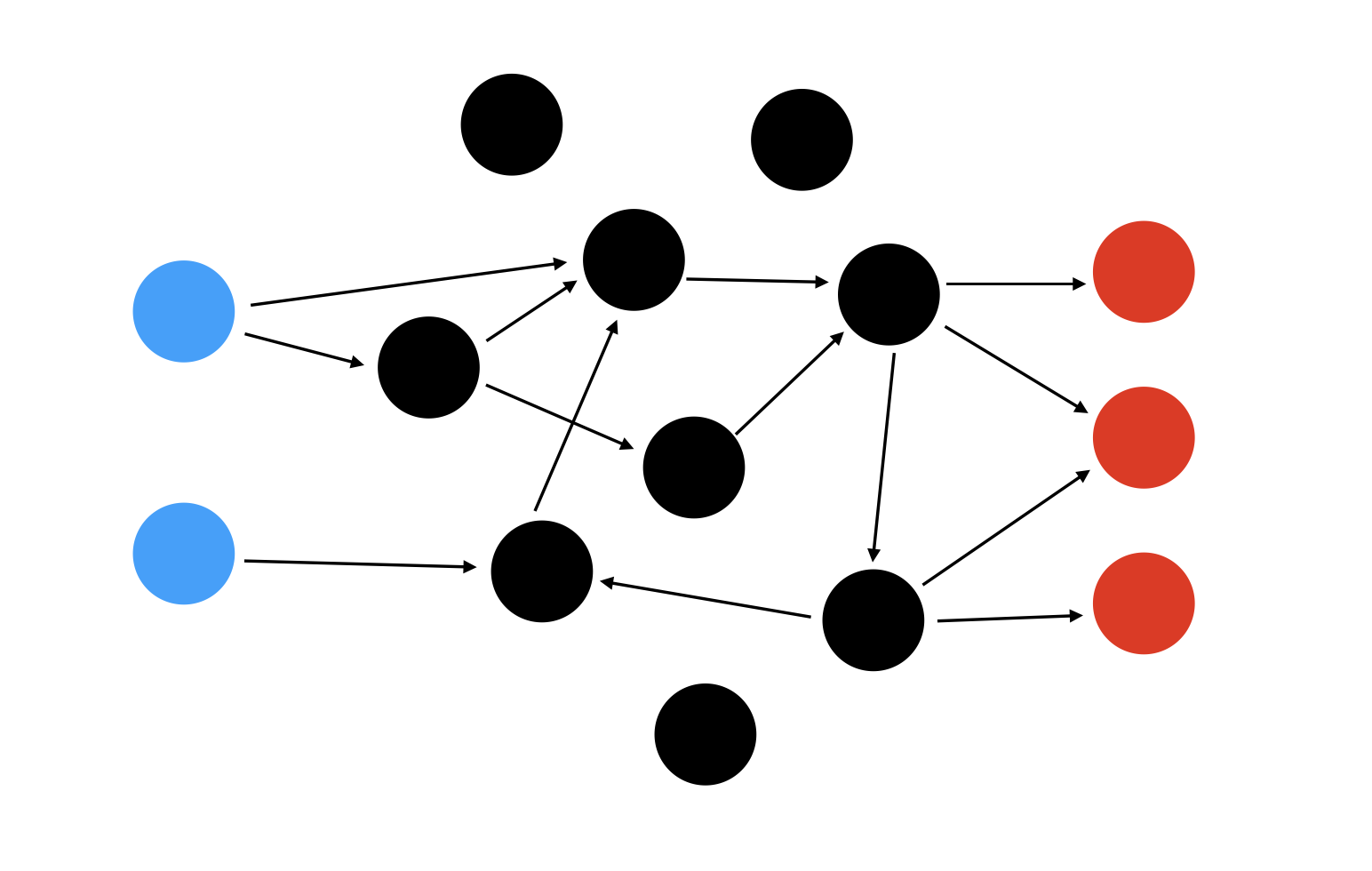}}
    \caption{LNDP components. \textbf{(a) Node model} Node features $h^t$, activations $v^t$ as well as additional graph structural features are fed through a single Graph Transformer layer whose output is fed as input to a GRU to obtain new nodes states $h^{t+1}$. \textbf{(b) Edge model} Edges are also modelled with GRUs and take as input pre and post-synaptics nodes' states and the last reward received $r^t$. \textbf{(c) Network topology} Networks are divided in input (blue), hidden (black) and output (red) neurons. Connections can only exist from input to hidden, hidden to hidden, and hidden to output. Some nodes can have no connections at all and the total number of hidden nodes is constant. Hyperparameters $\mu^{conn}$ and $\sigma^{conn}$ define the distribution (truncated normal) of the initial network density.}
    \label{fig:model}
\end{figure}

We describe here an instance of LNDPs based on Graph Transformer layers (GT) \citep{dwivedi2021} for the node update. More precisely, nodes are updated using Gated Recurrent Units (GRUs) \citep{chung2014} and take as input the output of a GT layer as depicted in Figure~\ref{fig:model}a, which enables nodes to ``perceive" the network and self-organize. The GT layer takes as input the concatenation of nodes activation $v^t$, nodes' states $h^t$, and structural graph features associated with the nodes. Structural features are the in-degree, out-degree, and total degree, as well as a one-hot encoding indicating if the node is an input, hidden, or output node. Moreover, we introduce edge features in the attention layer as proposed in \citet{dwivedi2021}. We also augment edge features with structural features which are 2 bits indicating if there is a forward or backward connection between nodes and a bit indicating if the edge is a self-loop. Intuitively, the graph transformer layer is used to model node-node interactions allowing them to self-organize and differentiate. We chose the GT rather than another graph neural network architecture as it seems easier to train. Importantly, it allowed to keep nodes' states from collapsing (i.e.\ all becoming the same), a recurrent problem with NDPs \citep{pedersen2024}. Node embeddings outputted by the GT are then fed into the GRU unit to update nodes' states ($h_i^{t+1} = GRU(h^t_i, x_i^t)$ where $x_i^t$ is the output of the GT layer). Importantly, all nodes share the same GRU parameters.
Synapses are also modeled as GRUs with shared parameters taking as input the concatenation of the pre and post-synaptic neurons' states as well as their activity and the reward received at the last time step $r$ as depicted in Figure~\ref{fig:model}b.

Pruning and synaptogenesis functions ($f_{\theta}^-$ and $f_{\theta}^+$ respectively) are modeled with multi-layer perceptrons (MLPs) with 1 hidden layer and ReLU activation function followed by a sigmoid to output the probability of respective actions. In this work, the dynamical model is defined as follows:
\begin{equation} \label{eq:dyn_rec}
    \phi(v^t, w^t, o^t) = tanh(\hat{v}^t \cdot w^t), 
\end{equation}
where $\hat{v}^t$ is the the activation vector with input nodes'  activations clamped to the values of the observation $o^t$. The weight matrix $w^t$ is obtained by taking the first element of the edges states ($w^t_{ij} = e^t_{ij,0}$). 

The initial network topology (i.e.\ adjacency matrix, see Figure~\ref{fig:model}c) is randomly initialized according to connection parameters $\mu_{conn}$ and standard deviation $\sigma_{conn}$, which together define a truncated normal distribution from which the connection probability is sampled: $\mathbb{P}(A^0_{ij}=1) \sim \mathcal{N}^{[0,1]}(\mu^{conn}, \sigma^{conn})$. The connection probability defines the probability that an edge exists from input to hidden, hidden to hidden, and hidden to output neurons at time $t=0$. There is a fixed number of nodes $N$ but some nodes can have no connections at all. Nodes' activations are initialized as zeros. Nodes' and edges' initial features ($h^0$ and $e^0$ respectively) are sampled from a uniform distribution between $-1$ and $1$. 

While the model can differentiate between input, hidden, and output neurons as this information is given to the GT layer, it cannot know \textit{a priori} which of its input (output) nodes corresponds to which information (action). This is because neurons are initialized following the same distribution and have the same connection probabilities and thus the same expected connectivity degree. Thus, for the model to solve a task, it has to differentiate its inputs and output neurons in order to make the right connections. This can be achieved for input neurons by looking at specific activation patterns, either from the environment or from spontaneous activity (see next paragraph). For output neurons, this can be achieved by finding specific dynamical relations between output neurons, rewards and/or input neuron activations to infer the effect of each of the output neurons. 

\subsection{Spontaneous activity} \label{sec:pre-exp-act}

We also study a mechanism for pre-experience (i.e.\ before receiving feedback from the environment) development of networks based on spontaneous activity, which has been shown to help the self-organization and learning of convolutional neural networks \citep{raghavan2019a}. In our case, we implement spontaneous activity through a learnable generative model generating spontaneous activity of the input nodes replacing observations from the environment, which we denote as $o_{SA}^t$. While many different models could be used, we chose to train a Ornstein-Uhlenbeck stochastic process defined as: 

\begin{equation} 
    \begin{cases}
        o_{SA}^{t+1} = o_{SA}^t + \alpha(\mu-o_{SA}^t) + W^t,\\ 
        W^t \sim \mathcal{N}(0, \Sigma).
    \end{cases}
\end{equation}
    
Mean $\mu \in \Real^{do}$, covariance $\Sigma \in \Real^{do\times do}$, and time constants $\alpha \in \Real^{do}$ are learnable parameters of the model where $do$ is the observation space dimensionality. This phase takes place before the environmental phase and lasts for $T^{SA}$ steps. Using spontaneous activity for pre-experience development enables the use of the same synaptic plasticity mechanisms as in the environmental phase.

\begin{figure*}[ht!]
    \centering
    \includegraphics[width=\textwidth]{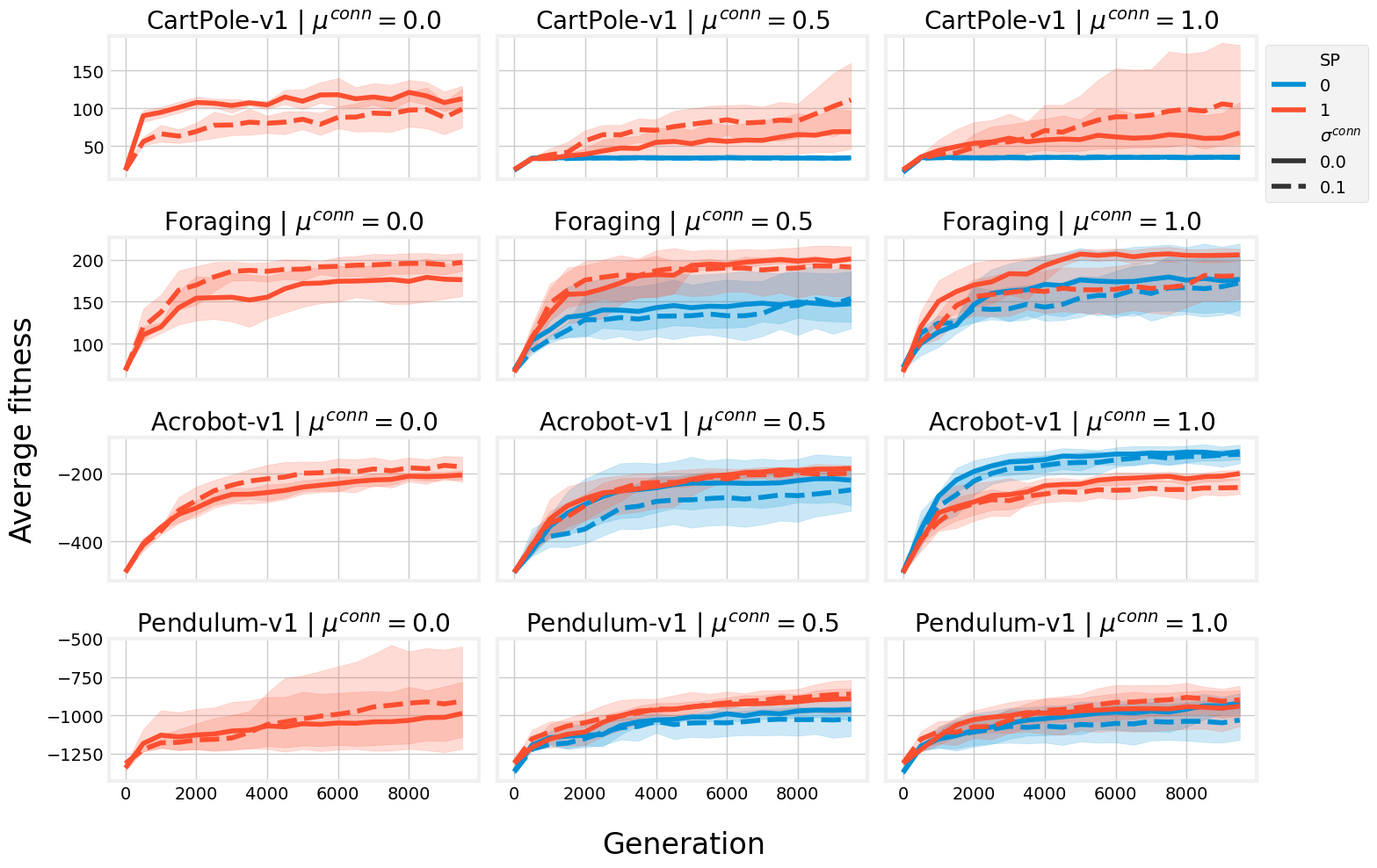}
    \caption{Training curves of the LNDP with varying initialization distributions (where $\mu^{conn}$ is the mean connection probability and $\sigma^{conn}$ its variance) and structural plasticity (SP)  enabled (red) and disabled (blue). For all conditions, structurally plastic LNDPs outperform non structurally plastic ones in Cartpole and Foraging.  Models without PS are not evaluated on empty networks ($\mu^{conn}=0$) as they would obviously fail (the network would remain empty).}
    \label{fig:results}

\end{figure*}

\section{Methods}

We optimized LNDPs on a set of reinforcement learning tasks using the Covariance Matrix Adaptation Evolutionary strategy (CMA-ES). The following sections give more details about the environments in use and the training procedure. LNDP hyperparameters used in this work are listed in Table~\ref{tab:params}. The code used to produce the results is available at: \git.

\begin{table}[htb!]
    \centering
    \subfloat[LNDP hyperparameters]{
        \begin{tabular}{|c|c|c|}
            \hline 
            $N$ & \# nodes &  $\in \{32, 64\}$\\
            \hline
            $dh$ & node features & $8$ \\
            \hline
            $de$ & edge features & $4$ \\
            \hline 
            $T^{SA}$ & developmental steps & $\in \{0, 100\}$\\
            \hline
            $\mu^{conn}$ & initial density average & $\in \{0, 0.5, 1.\}$\\
            \hline
            $\sigma^{conn}$ & initial density variance & $\in \{0, 0.1\}$\\
            \hline
        \end{tabular}
    }\\
    \subfloat[CMA-ES hyperparameters]{
        \begin{tabular}{|c|c|}
            \hline 
            Population size &  $128$\\
            \hline
            Initial CMA-ES variance  & $0.065$ \\
            \hline
            MC trials & $3$\\
            \hline
        \end{tabular}
    }
    \caption{Model and training hyperparameters.}
    \label{tab:params}
\end{table}

\subsection{Environments}\label{sec:envs}

We use three classical control tasks (Cartpole, Acrobot and Pendulum) and one foraging task with non-stationary dynamics requiring lifetime adaptivity from the agent. For all tasks, the fitness of an agent is defined as the average return of the agent over three different trials (i.e.\ different random seeds).

\begin{itemize}
    \item \textbf{Cartpole:} discrete control task in which the agent controls a cart with a pole on top. The agent has to balance the pole to keep it straight. This task features a 4-dimensional input space and a 2-dimensional discrete action space. The action is determined by the highest activated output nodes. The optimal reward is $500$. Importantly, it is difficult for an agent to find a solution directly solving the environment when starting from a random network (as is the case with LNDPs) as the agent needs to go from a random to a functional network in a short amount of time. Thus, we repeat the episodes without resetting the network states to give time for the model to adapt.
    \item \textbf{Acrobot:} discrete control task where the agent has to swing up a double pendulum above a certain height by applying torque to the joint, with the pendulum initially hanging downward. This environment features a 6-dimensional input space and a 3-dimensional discrete output space.
    \item \textbf{Pendulum:} continuous control task where the agent has to balance a pendulum in an upright position starting from a downward position. The observation space is 3-dimensional and there is a single continuous output.
    \item \textbf{Foraging:} a simple 1D grid with 5 cells and discrete 3-dimensional action space corresponding to moving right, left, and staying still. The agent starts in the center cell and can only observe which cell it is currently in. A food source is randomly placed at one end of the grid. When the agent reaches the food, it receives a reward of 10, and its position is reset. If the agent has not found the food after 10 steps, the environment is also reset. Each time the environment is reset there is a probability $p_{switch}$ (set to $0.5$ if not stated otherwise) that the food location is changed, making this environment non-stationary. 
\end{itemize}

\subsection{Training}

We optimize the model parameters on the different tasks using evosax \citep{lange2022a} implementation of the Covariance Matrix Adaptation Evolutionary Strategy (CMA-ES) \citep{hansen2023} with a population size of $128$ for all tasks except for the foraging environment, for which we use a population size of $256$. Fitness is obtained by averaging over three independent Monte-Carlo trials. Models are optimized for $10,000$ generations, with all presented results averaged over 5 different runs with different random seeds. Except for the foraging task, we use the gymnax library implementation of the environments \citep{gymnax2022github}. We compare the performance of the same model under different hyperparameters sets (see Table~\ref{tab:params}a) and with/without structural plasticity. CMA-ES hyperparamters can be found in Table~\ref{tab:params}b.

\begin{figure}[ht!]
    \centering
    \includegraphics[width=\columnwidth]{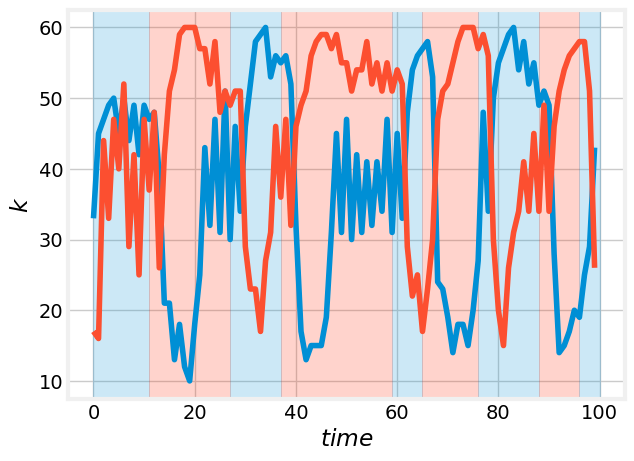}
    \caption{\textbf{Structural Plasticity.} Changes during the agent lifetime of output nodes' in-degrees in the foraging task. Red and blue lines correspond to left and right options respectively while shaded regions indicate the current location of the reward (red for left and blue for right). We can see that new connections are created towards the current best option while synapses towards the other action are pruned.}
    \label{fig:foraging}
\end{figure}

\section{Results}

\begin{figure*}[ht!]
    \centering
    \includegraphics[width=\textwidth]{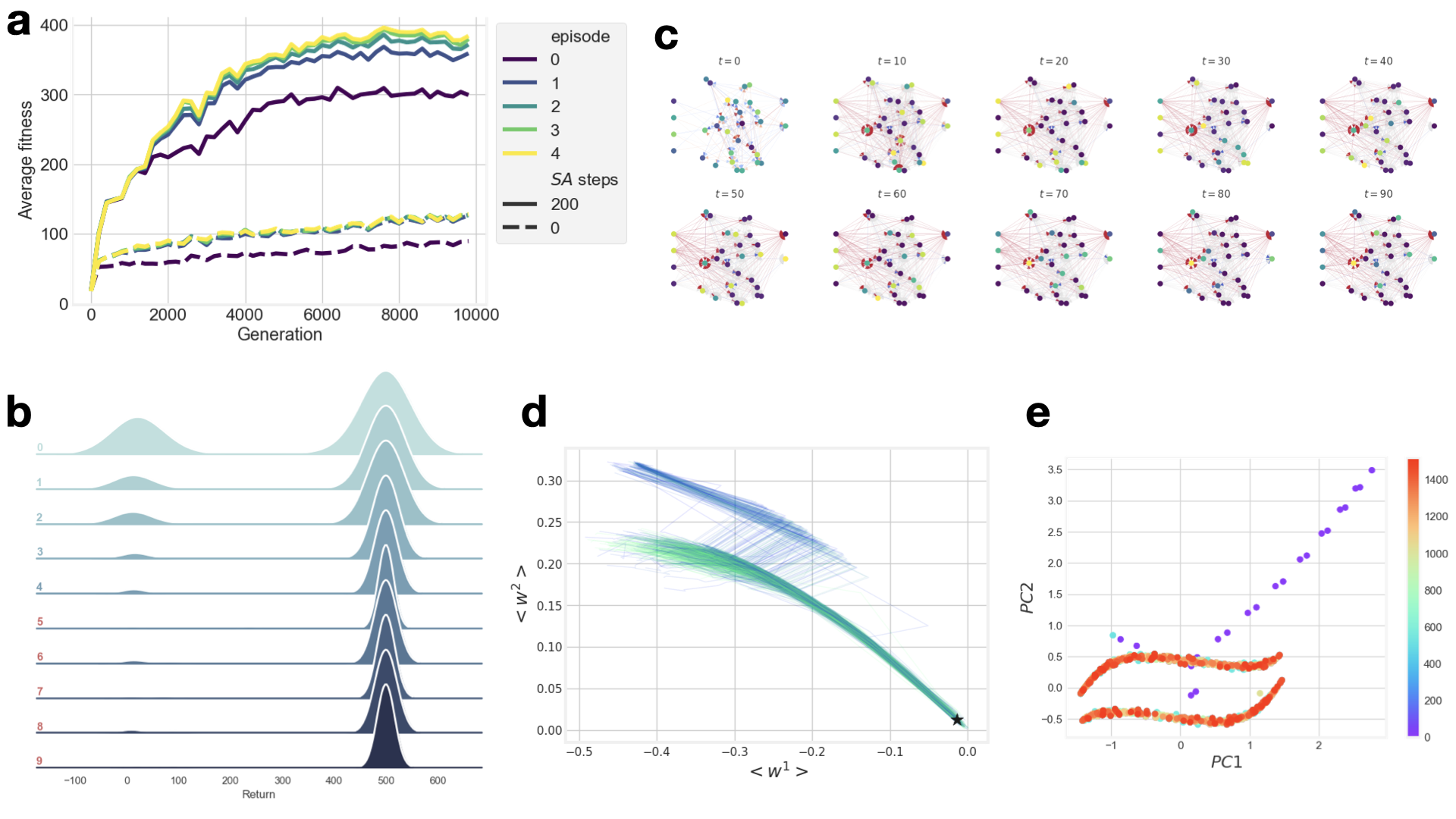}
    \caption{\textbf{LNDPs in CartPole.} (\textbf{a})  Training curves showing average reward of the population for each episode separately in CartPole. (\textbf{b}) Distribution of rewards at each episode (from top to bottom) obtained by one evolved LNDP trained on 5 episodes, episodes 5 to 9 (shown in red) are thus out of the training distribution. (\textbf{c}) Evolution of a network through its SA phase. Node colors code for the states $h^t$ and edge colors for weights $w^t$. Inputs are on the left and outputs on the right. (\textbf{d}) 1,000 developmental trajectories of a single evolved LNDP in the weights distribution space formed by the first two statistical moments (mean and variance of the weights). Trajectories correspond to different random initializations and colors code for the agent return (bluer lines correspond to failures and greener to successes), with the $\star$  indicating the starting point. (\textbf{e}) Trajectory of the network activations $v^t$ through time (dimensions are reduced through PCA).}
    \label{fig:sa}
\end{figure*}

Results from the evolutionary runs are shown in Figure~\ref{fig:results}. Solutions were discovered in all of the tested environments. We found differences between SP models and non-SP ones in the Cartpole environment, which represent a particularly difficult challenge: discovering sample efficient learning algorithms from very few samples. In our setting, the number of steps the agent experiences --  and thus the number of update steps of the model -- directly depends on its skills as the episode immediately ends after the pole has lost its balance. Consequently, a bad agent will see much fewer steps than a good agent. This creates a hard optimization problem at the evolutionary scale where an agent needs high sample efficiency for fast adaptation and this sample efficiency must itself be discovered from very few samples. Interestingly, while models without SP completely fail here, model endowed with SP are able to find well-performing solutions { (it should be noted that while the average return over the population does not reach the highest reward, i.e.\ 500, the maximum fitness per generation does)}. 

When looking at the rewards obtained at each episode we observe that SP models starting from empty networks reach higher rewards at the first episode and higher overall fitness showing fast adaptation capabilities as networks are able to self-organize into functional networks before the pole has lost its balance (for indication, a random policy would lose the balance in around $20$ steps). While no evolved LNDP is able to always find a functional network in the first episode, they are able to reach such a solution using lifetime learning. It should be noted that models with SP have more parameters to optimize ($1,854$ vs $1,468$ for non-SP models) and also have a much larger output space as they describe dynamics over weights and topologies. 

In the foraging environment with non-stationary rewards, we found that SP models consistently reached higher average population fitness than non-SP models while both reached similar maximum fitness. It should be noted that looking at average fitness is more informative as maximum fitness can easily be reached by chance. These results indicate that SP promotes better adaptivity in non-stationary settings. Most of the found solutions go on the side where the reward has been last obtained and go back to the other side if it has not been found, which is an optimal strategy when the switching probability is inferior or equal to $50\%$. Interestingly, these changes can be implemented as weight changes or structural changes where excitatory connections are created between hidden nodes and the output node corresponding to the new best option as shown in Figure~\ref{fig:foraging}.

No significant differences have been found in Acrobot except in the case of fully connected networks ($\mu^{conn}=1$) where models without SP reached higher average fitness than models with SP.
Finally, the pendulum environment, which is the only continuous control problem, appeared as the most challenging and the only one for which solutions where found only in a few runs and uniquely for SP models starting from empty networks (i.e.\ $\mu^{conn}=0$) with initialization variance $\sigma^{conn}=0.1$. 

\begin{figure}[hbtp]
    \centering
    \includegraphics[width=\columnwidth]{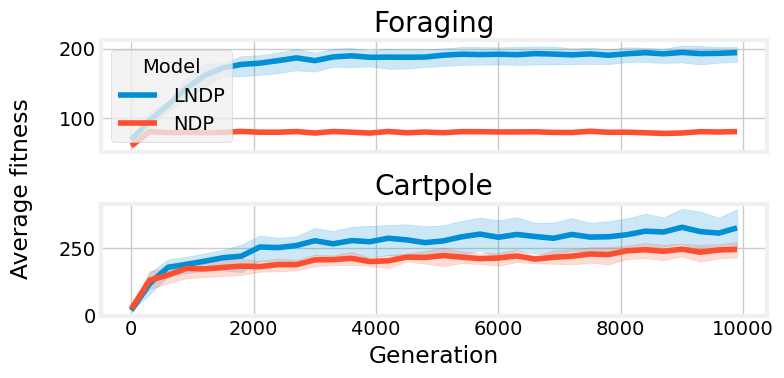}
    \caption{\textbf{Performance of LNDP and NDP in Foraging and Cartpole environments.} The NDP is obtained by ablating the network updates during lifetime (i.e.\ after the SA phase). Both models go through $100$ SA steps. While the performance of both approaches is more similar in the CartPole domain, the Foraging task requires an agent to adapt during its lifetime, which only the LNDP is capable of.}
    \label{fig:ndp}
\end{figure}

In the CartPole environment, it is particularly difficult for models without SA to reach good performance in the first episode as the domain requires fast adaptation for the pole to not directly fall over. Models with SA, on the other hand, display innate skills in solving the task at the first episode, demonstrating the ability of the model to reach target functional networks in a reward-independent and self-organized manner (Figure~\ref{fig:sa}). We also observe that lifetime learning appears in later generations during training for models with SA, the first $\sim 2,000$ generations being characterized by an increase in innate skills only (Figure~\ref{fig:sa}a). This could be explained by reduced pressure on learning abilities. When training models without SA for much longer (i.e.\ $50 \times 10^3$ generations), models were found that can adapt quickly enough to succeed in the first episode, showing increased sample efficiency. Furthermore, we observed that while LNDPs without a SA phase tend to not improve further after the second episode, models trained with  SA  display stronger differences in between all episodes (Figure~\ref{fig:sa}a). The latter observation could indicate synergies between lifetime learning and SA-driven development. We also compared results with an NDP, i.e.\ an LNDP not able to modify its state during its lifetime. Results shown in Figure~\ref{fig:ndp} demonstrate that the LNDP outperforms the NDP in Foraging and CartPole. In the Foraging environment, the NDP is not able to adapt at all which it could only achieve by keeping track of environmental states in the activation patterns.  

While we observed a high degree of variability across runs, in terms of dynamics and operating mechanisms, some features were commonly observed. First, networks tend to display short critical periods during which most of the changes happen. Interestingly, for some models, we observed that failures in the task were associated with abnormal bifurcations in the developmental trajectory happening during this critical period as shown in Figure~\ref{fig:sa}(d). Having critical periods after which almost no change happens could represent a trade-off between adaptivity and stability. In particular, we observed that LNDPs evolved in the CartPole environment both with and without SA were very consistent through time, even in episodes unseen during training (see Figure~\ref{fig:sa}b  where the agent is trained on 5 episodes and evaluated on 10) were some solutions keep improving. 

Drawing general conclusions about the discovered learning mechanisms is difficult because of the high variability between seeds. While we have not observed the emergence of noticeable structures in the grown networks, most of them display less than fully-connected structures (Average network densities at the end of training: Cartpole: $0.79 \pm 0.23$, Foraging: $0.72 \pm 0.16$, Acrobot: $0.69 \pm 0.22$, Pendulum: $0.24 \pm 0.36$, densities are corrected to account for impossible connections). Interestingly, we found that failed training runs, mostly observed in the Pendulum environment, are characterized by a fast convergence of the population towards producing empty networks, which explains the lower average density found for Pendulum).  

\section{Discussion}

Extending previous approaches on NDPs \citep{najarro2023} to lifetime experience-dependant learning, we showed that LNDPs can be evolved to drive the self-organization of ANNs into functional networks able to solve control tasks with various properties. While structural plasticity is a poorly addressed property in the context of artificial neural networks, we showed evidence that it enables LNDPs to display better adaptivity in  (1) an environment with non-stationary dynamics requiring continuous adaptation, and (2) in an environment requiring fast adaptation (with the additional difficulty of having to learn this model from only a few experiences). 

Spontaneous activity (SA) driven development, while being an important mechanism in biological neural networks \citep{luhmann2016,leighton2016,kirkby2013a},  has not been well explored in ANNs (see \citet{raghavan2019a} for an example). We showed in this work that SA, modeled as a simple stochastic process generating input neuron activity, can be effectively used in concert with LNDPs to drive the network's self-organization in a pre-environmental phase allowing agents to display innate skills and better adaptivity. Additionally, no extra component is needed during the developmental phase; in other words, the same components drive the network's self-organization regardless of whether inputs come from SA or the environment. This property enabled us to uncover synergies between the two phases.  

LNDPs represent a promising path toward self-assembling neural networks able to adapt quickly. However, work remains to achieve performances comparable to architectures traditionally used in the context of reinforcement learning. While we have shown that LNDPs can be trained to solve and implement learning algorithms in simple environments, future research must determine how they scale to more complex ones with higher dimensions. As instances of complex self-organized systems, their training represents a challenging optimization problem characterized by many interactions over long periods of time.

While the architecture presented in this paper is intentionally general, using versatile components such as GRUs and Graph Transformer layers, and introducing very few inductive biases, future works should explore how to efficiently constrain the LNDPs output space while keeping the possibility of discovering unanticipated models. Inspiration could be drawn from developmental neurosciences to introduce useful biases in learning and developmental mechanisms. For instance, different synaptic plasticity rules have been recently proposed leading to the emergence of properties usually found in biological neural networks \citep{lynn2024a,achterberg2023}. The general form of these rules could be used to guide the learning of LNDPs.

Moreover, while we used classical evolutionary strategies together with standard training methods, we believe that more focus should be given to this aspect for more efficient training. Methods such as novelty search \citep{lehman2011c}, quality diversity \citep{pugh2016a} or intrinsically motivated optimization \citep{laversanne-finot2021} could represent particularly relevant approaches for the problem at hand. Graph generation, which is in our case an indirect task of the model, is a notoriously hard problem necessitating large models and large amounts of data in a supervised setting with much more informative feedback \citep{zhu2022a}. Moreover, we found that training the wiring rules and the synaptic plasticity model made the optimization particularly difficult because of the strong dependencies between the two mechanisms.

Overall, we believe that by accounting for both pre-experience learning through SA-driven development and experience-dependent adaptation, LNDPs represent a promising path toward new types of self-organizing ANNs that can narrow the gap between natural and artificial adaptivity in the future.
\section{Acknowledgements}
This project was supported by a European Research Council (ERC) grant (GA no. 101045094, project ”GROW-AI”).

\footnotesize
\bibliographystyle{apalike}
\bibliography{PNDP} 

\begin{thebibliography}{}

\bibitem[Achterberg et~al., 2023]{achterberg2023}
Achterberg, J., Akarca, D., Strouse, D.~J., Duncan, J., and Astle, D.~E. (2023).
\newblock Spatially embedded recurrent neural networks reveal widespread links between structural and functional neuroscience findings.
\newblock {\em Nature Machine Intelligence}, 5(12):1369--1381.

\bibitem[Andrychowicz et~al., 2016]{andrychowicz2016}
Andrychowicz, M., Denil, M., G{\'o}mez, S., Hoffman, M.~W., Pfau, D., Schaul, T., Shillingford, B., and {de Freitas}, N. (2016).
\newblock Learning to learn by gradient descent by gradient descent.
\newblock In {\em Advances in {{Neural Information Processing Systems}}}, volume~29. Curran Associates, Inc.

\bibitem[Aswolinskiy and Pipa, 2015]{aswolinskiy2015}
Aswolinskiy, W. and Pipa, G. (2015).
\newblock {{RM-SORN}}: A reward-modulated self-organizing recurrent neural network.
\newblock {\em Frontiers in Computational Neuroscience}, 9.

\bibitem[Bednar and Miikkulainen, 1998]{bednar1998}
Bednar, J.~A. and Miikkulainen, R. (1998).
\newblock Pattern-{{Generator-Driven Development}} in {{Self-Organizing Models}}.
\newblock In Bower, J.~M., editor, {\em Computational {{Neuroscience}}: {{Trends}} in {{Research}}, 1998}, pages 317--323. Springer US, Boston, MA.

\bibitem[Bhattasali et~al., 2022]{bhattasali2022a}
Bhattasali, N., Zador, A.~M., and Engel, T. (2022).
\newblock Neural {{Circuit Architectural Priors}} for {{Embodied Control}}.
\newblock {\em Advances in Neural Information Processing Systems}, 35:12744--12759.

\bibitem[Caroni et~al., 2012]{caroni2012}
Caroni, P., Donato, F., and Muller, D. (2012).
\newblock Structural plasticity upon learning: Regulation and functions.
\newblock {\em Nature Reviews Neuroscience}, 13(7):478--490.

\bibitem[Chung et~al., 2014]{chung2014}
Chung, J., Gulcehre, C., Cho, K., and Bengio, Y. (2014).
\newblock Empirical {{Evaluation}} of {{Gated Recurrent Neural Networks}} on {{Sequence Modeling}}.

\bibitem[de~Campos et~al., 2011]{campos2011}
de~Campos, L. M.~L., Roisenberg, M., and de~Oliveira, R. C.~L. (2011).
\newblock Automatic design of {{Neural Networks}} with {{L-Systems}} and genetic algorithms - {{A}} biologically inspired methodology.
\newblock In {\em The 2011 {{International Joint Conference}} on {{Neural Networks}}}, pages 1199--1206.

\bibitem[Dwivedi and Bresson, 2021]{dwivedi2021}
Dwivedi, V.~P. and Bresson, X. (2021).
\newblock A {{Generalization}} of {{Transformer Networks}} to {{Graphs}}.

\bibitem[Ferigo and Iacca, 2023]{ferigo2023}
Ferigo, A. and Iacca, G. (2023).
\newblock Self-building neural networks.
\newblock In {\em Proceedings of the Companion Conference on Genetic and Evolutionary Computation}, pages 643--646.

\bibitem[Ginsburg and Jablonka, 2021]{ginsburg2021}
Ginsburg, S. and Jablonka, E. (2021).
\newblock Evolutionary transitions in learning and cognition.
\newblock {\em Philosophical Transactions of the Royal Society B: Biological Sciences}, 376(1821):20190766.

\bibitem[Ha et~al., 2022]{ha2022b}
Ha, D., Dai, A.~M., and Le, Q.~V. (2022).
\newblock {{HyperNetworks}}.
\newblock In {\em International {{Conference}} on {{Learning Representations}}}.

\bibitem[Ha and Tang, 2022]{ha2022a}
Ha, D. and Tang, Y. (2022).
\newblock Collective intelligence for deep learning: {{A}} survey of recent developments.
\newblock {\em Collective Intelligence}, 1(1):26339137221114874.

\bibitem[Hansen, 2023]{hansen2023}
Hansen, N. (2023).
\newblock The {{CMA Evolution Strategy}}: {{A Tutorial}}.

\bibitem[Hochreiter et~al., 2001]{hochreiter2001}
Hochreiter, S., Younger, A.~S., and Conwell, P.~R. (2001).
\newblock Learning to {{Learn Using Gradient Descent}}.
\newblock In Dorffner, G., Bischof, H., and Hornik, K., editors, {\em Artificial {{Neural Networks}} --- {{ICANN}} 2001}, pages 87--94, Berlin, Heidelberg. Springer.

\bibitem[Kirkby et~al., 2013]{kirkby2013a}
Kirkby, L.~A., Sack, G.~S., Firl, A., and Feller, M.~B. (2013).
\newblock A {{Role}} for {{Correlated Spontaneous Activity}} in the {{Assembly}} of {{Neural Circuits}}.
\newblock {\em Neuron}, 80(5):1129--1144.

\bibitem[Kirsch and Schmidhuber, 2022]{kirsch2022}
Kirsch, L. and Schmidhuber, J. (2022).
\newblock Meta {{Learning Backpropagation And Improving It}}.

\bibitem[Kudithipudi et~al., 2022]{kudithipudi2022}
Kudithipudi, D., {Aguilar-Simon}, M., Babb, J., Bazhenov, M., Blackiston, D., Bongard, J., Brna, A.~P., Chakravarthi~Raja, S., Cheney, N., Clune, J., Daram, A., Fusi, S., Helfer, P., Kay, L., Ketz, N., Kira, Z., Kolouri, S., Krichmar, J.~L., Kriegman, S., Levin, M., Madireddy, S., Manicka, S., Marjaninejad, A., McNaughton, B., Miikkulainen, R., Navratilova, Z., Pandit, T., Parker, A., Pilly, P.~K., Risi, S., Sejnowski, T.~J., Soltoggio, A., Soures, N., Tolias, A.~S., {Urbina-Mel{\'e}ndez}, D., {Valero-Cuevas}, F.~J., {van de Ven}, G.~M., Vogelstein, J.~T., Wang, F., Weiss, R., {Yanguas-Gil}, A., Zou, X., and Siegelmann, H. (2022).
\newblock Biological underpinnings for lifelong learning machines.
\newblock {\em Nature Machine Intelligence}, 4(3):196--210.

\bibitem[Lange, 2022a]{lange2022a}
Lange, R.~T. (2022a).
\newblock Evosax: {{JAX-based Evolution Strategies}}.

\bibitem[Lange, 2022b]{gymnax2022github}
Lange, R.~T. (2022b).
\newblock {{gymnax}}: {{A JAX-based}} reinforcement learning environment library.

\bibitem[{Laversanne-Finot} et~al., 2021]{laversanne-finot2021}
{Laversanne-Finot}, A., P{\'e}r{\'e}, A., and Oudeyer, P.-Y. (2021).
\newblock Intrinsically {{Motivated Exploration}} of {{Learned Goal Spaces}}.
\newblock {\em Frontiers in Neurorobotics}, 14.

\bibitem[Lazar et~al., 2009]{lazar2009}
Lazar, A., Pipa, G., and Triesch, J. (2009).
\newblock {{SORN}}: {{A Self-Organizing Recurrent Neural Network}}.
\newblock {\em Frontiers in Computational Neuroscience}, 3.

\bibitem[Lehman and Stanley, 2011]{lehman2011c}
Lehman, J. and Stanley, K.~O. (2011).
\newblock Novelty {{Search}} and the {{Problem}} with {{Objectives}}.
\newblock In Riolo, R., Vladislavleva, E., and Moore, J.~H., editors, {\em Genetic {{Programming Theory}} and {{Practice IX}}}, pages 37--56. Springer New York, New York, NY.

\bibitem[Leighton and Lohmann, 2016]{leighton2016}
Leighton, A.~H. and Lohmann, C. (2016).
\newblock The {{Wiring}} of {{Developing Sensory Circuits}}---{{From Patterned Spontaneous Activity}} to {{Synaptic Plasticity Mechanisms}}.
\newblock {\em Frontiers in Neural Circuits}, 10.

\bibitem[Lowell and Pollack, 2016]{lowell2016}
Lowell, J. and Pollack, J. (2016).
\newblock Developmental {{Encodings Promote}} the {{Emergence}} of {{Hierarchical Modularity}}.
\newblock In {\em {{ALIFE}} 2016, the {{Fifteenth International Conference}} on the {{Synthesis}} and {{Simulation}} of {{Living Systems}}}, pages 344--351. MIT Press.

\bibitem[Luhmann et~al., 2016]{luhmann2016}
Luhmann, H.~J., Sinning, A., Yang, J.-W., {Reyes-Puerta}, V., St{\"u}ttgen, M.~C., Kirischuk, S., and Kilb, W. (2016).
\newblock Spontaneous {{Neuronal Activity}} in {{Developing Neocortical Networks}}: {{From Single Cells}} to {{Large-Scale Interactions}}.
\newblock {\em Frontiers in Neural Circuits}, 10.

\bibitem[Lynn et~al., 2024]{lynn2024a}
Lynn, C.~W., Holmes, C.~M., and Palmer, S.~E. (2024).
\newblock Heavy-tailed neuronal connectivity arises from {{Hebbian}} self-organization.
\newblock {\em Nature Physics}, 20(3):484--491.

\bibitem[May, 2011]{may2011}
May, A. (2011).
\newblock Experience-dependent structural plasticity in the adult human brain.
\newblock {\em Trends in Cognitive Sciences}, 15(10):475--482.

\bibitem[Najarro and Risi, 2020]{najarro2020}
Najarro, E. and Risi, S. (2020).
\newblock Meta-{{Learning}} through {{Hebbian Plasticity}} in {{Random Networks}}.
\newblock In {\em Advances in {{Neural Information Processing Systems}}}, volume~33, pages 20719--20731. Curran Associates, Inc.

\bibitem[Najarro et~al., 2022]{najarro2022b}
Najarro, E., Sudhakaran, S., Glanois, C., and Risi, S. (2022).
\newblock {{HyperNCA}}: {{Growing Developmental Networks}} with {{Neural Cellular Automata}}.

\bibitem[Najarro et~al., 2023]{najarro2023}
Najarro, E., Sudhakaran, S., and Risi, S. (2023).
\newblock Towards {{Self-Assembling Artificial Neural Networks}} through {{Neural Developmental Programs}}.
\newblock In {\em {{ALIFE}} 2023: {{Ghost}} in the {{Machine}}: {{Proceedings}} of the 2023 {{Artificial Life Conference}}}. MIT Press.

\bibitem[Parisi et~al., 2019]{parisi2019}
Parisi, G.~I., Kemker, R., Part, J.~L., Kanan, C., and Wermter, S. (2019).
\newblock Continual lifelong learning with neural networks: {{A}} review.
\newblock {\em Neural Networks}, 113:54--71.

\bibitem[Pedersen et~al., 2024]{pedersen2024}
Pedersen, J.~W., Plantec, E., Nisioti, E., Montero, M., and Risi, S. (2024).
\newblock Structurally {{Flexible Neural Networks}}: {{Evolving}} the {{Building Blocks}} for {{General Agents}}.

\bibitem[Pugh et~al., 2016]{pugh2016a}
Pugh, J.~K., Soros, L.~B., and Stanley, K.~O. (2016).
\newblock Quality {{Diversity}}: {{A New Frontier}} for {{Evolutionary Computation}}.
\newblock {\em Frontiers in Robotics and AI}, 3.

\bibitem[Raghavan and Thomson, 2019]{raghavan2019a}
Raghavan, G. and Thomson, M. (2019).
\newblock Neural networks grown and self-organized by noise.
\newblock In {\em Advances in {{Neural Information Processing Systems}}}, volume~32. Curran Associates, Inc.

\bibitem[Risi, 2021]{risi2021selfassemblingAI}
Risi, S. (2021).
\newblock The future of artificial intelligence is self-organizing and self-assembling.
\newblock {\em sebastianrisi.com}.

\bibitem[Sandler et~al., 2021]{sandler2021b}
Sandler, M., Vladymyrov, M., Zhmoginov, A., Miller, N., Madams, T., Jackson, A., and Arcas, B. A.~Y. (2021).
\newblock Meta-{{Learning Bidirectional Update Rules}}.
\newblock In {\em Proceedings of the 38th {{International Conference}} on {{Machine Learning}}}, pages 9288--9300. PMLR.

\bibitem[Soltoggio et~al., 2018]{soltoggio2018a}
Soltoggio, A., Stanley, K.~O., and Risi, S. (2018).
\newblock Born to learn: {{The}} inspiration, progress, and future of evolved plastic artificial neural networks.
\newblock {\em Neural Networks}, 108:48--67.

\bibitem[Stanley and Miikkulainen, 2003]{stanley2003a}
Stanley, K.~O. and Miikkulainen, R. (2003).
\newblock A {{Taxonomy}} for {{Artificial Embryogeny}}.
\newblock {\em Artificial Life}, 9(2):93--130.

\bibitem[Valsalam et~al., 2007]{valsalam2007}
Valsalam, V.~K., Bednar, J.~A., and Miikkulainen, R. (2007).
\newblock Developing {{Complex Systems Using Evolved Pattern Generators}}.
\newblock {\em IEEE Transactions on Evolutionary Computation}, 11(2):181--198.

\bibitem[Wang et~al., 2018]{wang2018c}
Wang, J.~X., {Kurth-Nelson}, Z., Kumaran, D., Tirumala, D., Soyer, H., Leibo, J.~Z., Hassabis, D., and Botvinick, M. (2018).
\newblock Prefrontal cortex as a meta-reinforcement learning system.
\newblock {\em Nature Neuroscience}, 21(6):860--868.

\bibitem[Zhu et~al., 2022]{zhu2022a}
Zhu, Y., Du, Y., Wang, Y., Xu, Y., Zhang, J., Liu, Q., and Wu, S. (2022).
\newblock A {{Survey}} on {{Deep Graph Generation}}: {{Methods}} and {{Applications}}.
\newblock In {\em Proceedings of the {{First Learning}} on {{Graphs Conference}}}, pages 47:1--47:21. PMLR.

\end{thebibliography}

\end{document}